\documentclass[authoryear,preprint,review,1p]{elsarticle}
\usepackage{enumitem}
\usepackage{xcolor}
\usepackage{array}
\usepackage{float}
\usepackage[hidelinks]{hyperref}
\newcolumntype{C}[1]{>{\centering\arraybackslash}p{#1}}
\usepackage{amssymb}
\usepackage{caption}
\usepackage{lineno}

\begin{document}

\begin{frontmatter}

\title{SeeTree - A modular, open-source system for tree detection and orchard localization}

\author[inst1]{Jostan Brown}
\author[inst1]{Cindy Grimm}
\author[inst1]{Joseph R. Davidson}

\affiliation[inst1]{organization={Collaborative Robotics and Intelligent Systems (CoRIS) Institute},
            addressline={Oregon State University}, 
            city={Corvallis OR},
            postcode={97331}, 
            country={USA}}

\begin{abstract}
Accurate localization is an important functional requirement for precision orchard management. However, there are few off-the-shelf commercial solutions available to growers. In this paper, we present SeeTree, a modular, open source embedded system for tree trunk detection and orchard localization that is deployable on any vehicle. Building on our prior work on vision-based in-row localization using particle filters, SeeTree includes several new capabilities. First, it provides capacity for full orchard localization including out-of-row headland turning. Second, it includes the flexibility to integrate either visual, GNSS, or wheel odometry in the motion model. During field experiments in a commercial orchard, the system converged to the correct location 99\% of the time over 800 trials, even when starting with large uncertainty in the initial particle locations. When turning out of row, the system correctly tracked 99\% of the turns (860 trials representing 43 unique row changes). To help support adoption and future research and development, we make our dataset, design files, and source code freely available to the community. 
\end{abstract}

\begin{keyword}

Localization \sep particle filter \sep orchard \sep embedded system 

\end{keyword}

\end{frontmatter}


\section{Introduction}

\label{sec:Introduction}
Over the past few decades, there has been significant adoption of digital technologies in commercial broadacre agriculture~(\cite{PrecisionAg_2023}). A few high impact examples include accurate and affordable RTK-GNSS systems, equipment auto steering, and unmanned aerial vehicle (UAV)-mounted sensor platforms. Today, farmers can even purchase autonomous tractors. However, compared to the row crop industry, there has been less adoption of precision technologies in the tree fruit and nut industries~(\cite{PerceptionsofPrecisionAgricultureTechnologiesintheUSFreshAppleIndustry}). One challenge is the large variability in orchard designs from farm to farm. Another constraint is the lack of low-cost, commercially available off-the-shelf solutions for localization within the orchard, which is an important enabling technology for many precision management activities. Whether the goal is using historical data to apply selective fertilizer rates targeted to individual trees' nutritional needs or recording per-tree fruit yields on an autonomous robotic harvester, \textit{a basic functional requirement is knowing where you are in the orchard}.  

One of the primary technical challenges with localizing in orchards at \textit{the tree level} is that tall, dense canopies interfere with GNSS measurements. GNSS equipment can be elevated above the canopy to improve accuracy, but these setups are expensive and cumbersome. For techniques that do not rely entirely on GNSS, such as Simultaneous Localization and Mapping (SLAM), perceptual aliasing (i.e. different parts of the orchard appear visually similar) and the lack of unique landmarks are also challenges~(\cite{Underwood_JFR_2015}). Improving the performance of SLAM-based techniques for orchard localization is an active area of research~(\cite{Aguiar_Robotics_2020,Tang_IEEE_2024}). \cite{Guevara_SmartAgTech_2024} recently published work on scan matching techniques for orchard localization that used point clouds obtained from LiDAR. Compared to LiDAR sensors, cameras are a cheaper alternative and can provide detailed information about a scene (e.g. semantic information). Combined with advances in deep learning, vision-based navigation systems are becoming increasingly common. For example, \cite{Sivakumar-RSS-24} recently presented CropFollow++, a vision-based navigation system that uses learned semantic keypoints to navigate small ground robots under dense cover crop canopies (in this case row centering and following). 

Orchards are semi-structured environments with many repetitive, stable features, such as support posts and tree trunks. Tree trunks are useful features for a vision-based navigation system for several reasons. First, although trees may be planted with uniform spacing, small variations in growth direction and rate are reflected in small variations in positions and trunk diameters. These are (largely) static and do not change (much) from year to year. Second, while tree canopies are dense/cluttered, often overlap, and change significantly over the growing season, the trunks are  visually easy to detect and measure, even if partially occluded.  

In our previous work~(\cite{Brown_COMPAG_2024}), we presented a vision-based system for localizing on an orchard map at the per tree level. The camera was pointed perpendicular to the vehicle's direction of travel, imaging the lower portion of the tree. In brief, the system first segments a tree trunk in an RGB-D image and then estimates its width. Then, both the position of the segmentation mask in the image and the trunk width are used to calculate the particle weights of a particle filter. While extensive field trials with a ground robot showed accurate localization even with uncertainty in the robot's initial starting pose, the system had two notable limitations. First, the system could not localize outside of the row during headland turning when there were no visible trees. Second, the system relied on the robot's wheel odometry in the motion model, which limited its ability to be deployed on vehicles without such sensors (e.g. conventional tractors). 

In this paper, we present \textbf{SeeTree}, an updated version of our localization system with the following improvements:
\begin{itemize}
    \item Capacity for full orchard localization, including out-of-row headland navigation
    \item Flexibility to integrate either visual, GNSS, or wheel odometry in the motion model
    \item Updated software with a wireless interactive display and a tuning app for adjusting filter parameters in changing environments
\end{itemize}
We validated our new design during field experiments at a commercial apple orchard. Here we present the system's architecture (Sec.~\ref{sec:System architecture}), validation method (Sec.~\ref{sec:Validation method}), and experimental results (Sec.~\ref{sec:Results}). Our primary contributions are i) a modular, open-source\footnote{\url{https://github.com/Jostan86/pf_orchard_localization}} orchard localization system that can be deployed on any vehicle, human-operated or autonomous; and ii) a quantitative evaluation and comparison of localization performance for multiple types of vehicle odometries.

\section{System architecture}
\label{sec:System architecture}
SeeTree was designed as a flexible system that can be easily integrated into a variety of vehicles, accommodating different sensor configurations and operational needs. In this section we present the system's architecture. We start by describing the software application for visualizing and fine-tuning the tree trunk segmentation and width estimation pipeline. We then review the system's hardware design, including sensor selection, networking and communications, and power support. Finally, we describe our implementation of different types of odometry in the particle filter's motion model.  

\subsection{Trunk width estimation and analysis software}

The first enhancement to our localization system are improvements to the Python package for estimating tree trunk widths, available at \url{https://github.com/Jostan86/trunk_width_estimation}. The core process for trunk width estimation has remained largely the same as described in Section 3.1 of our original paper~(\cite{Brown_COMPAG_2024}), but several new features have been added to improve usability and overall functionality. Key improvements include:

\begin{itemize}
    \item \textbf{PyQt Tuning Application:} A PyQt-based application has been developed to help visualize the trunk width estimation process and facilitate parameter tuning for new datasets. See Sec.~\ref{subsec:TuningAp} for additional details.
    \item \textbf{Restructured Code Base:} The code has been reorganized to simplify the addition of new operations for image processing, particularly additional filtering checks. This modular structure makes it easier for developers and researchers to expand the system's capabilities.
    \item \textbf{Use Cases/Examples:} A Jupyter Notebook and Python script have been created to provide a thorough example of the software package. These resources help the user understand the overall trunk width estimation process and how to add custom operations or modify existing ones within the new architecture.
    \item \textbf{Enhanced VSCode Dev-Container Support:} The development environment has been improved with an updated VSCode dev-container and Dockerfile setup. This setup ensures that the package and example code is effectively turn-key runnable on Ubuntu systems, minimizing the complexity of environment configuration.
\end{itemize}

\subsubsection{PyQt Tuning Application}
\label{subsec:TuningAp}
The tuning application is designed to provide both visualization and parameter tuning capabilities to improve the robustness of the trunk width estimation system for a custom dataset. The app operates by loading a dataset of images containing tree trunks with known ground truth widths, allowing users to interactively adjust parameters for various operations in the system to optimize width estimation performance for the dataset. Examples of tunable parameters include the depth threshold used to filter out distant detections, the confidence threshold of the segmentation model, the margin used to identify segmentations near the image edge, the minimum number of valid depth points required to retain a detection, and the percentile value used to estimate trunk depth from the histogram of depth values.

\begin{figure}[h]
  \centering
  \includegraphics[width=.98\textwidth]{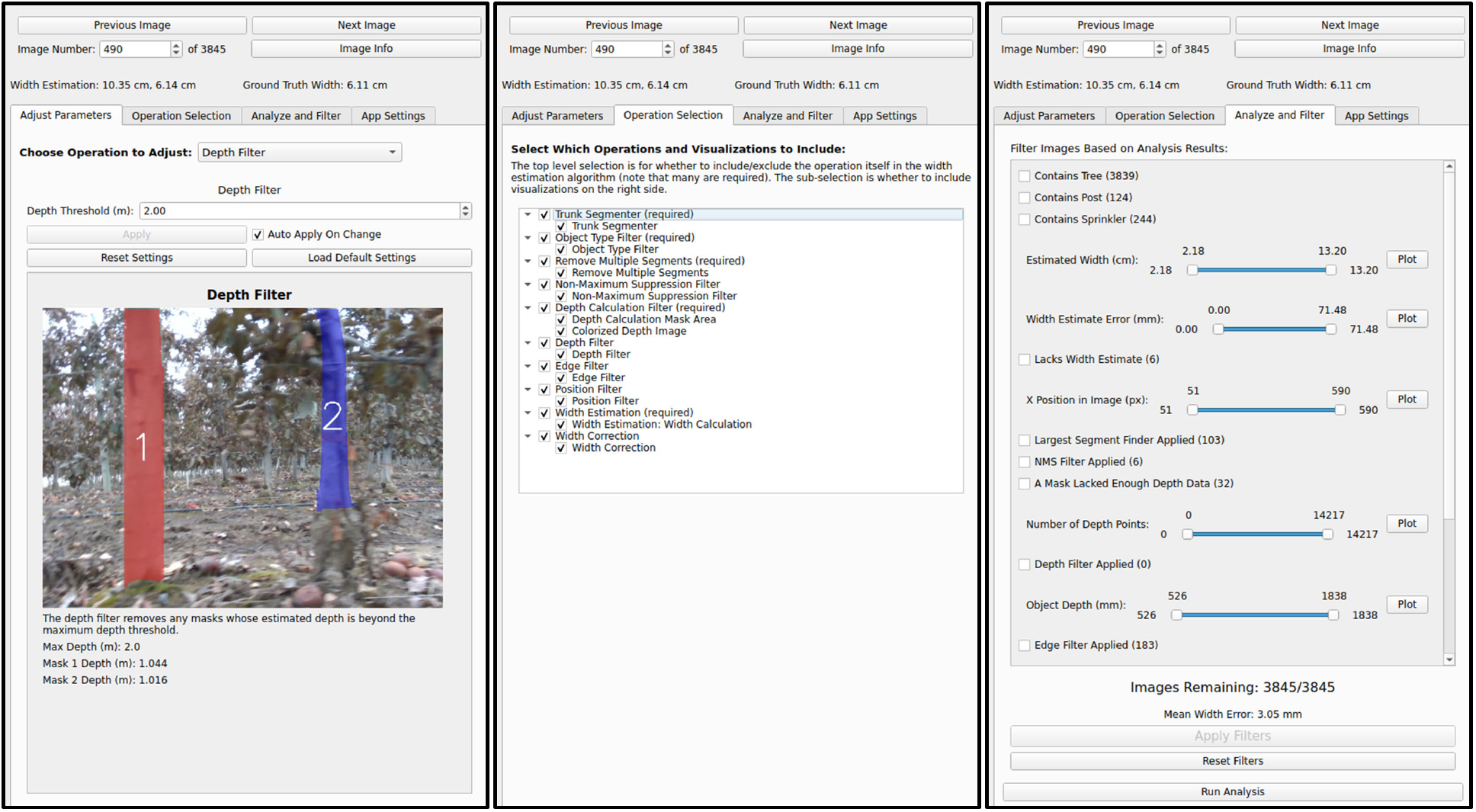}
  \caption{There are three sets of tools available in three tabs on the left side of the app for parameter tuning (left), operation/visualization toggling (middle), and dataset filtering (right).}
  \label{fig:width_app_figure}
\end{figure}

The application's interface is divided into two main sections. The right side displays visualizations of the trunk width estimation process for the current image, including explanations of each step, shown in Fig.~\ref{fig:operation_visualization}. The left side features three tabs for parameter tuning, operation control, and dataset filtering, as described below and shown in Fig.~\ref{fig:width_app_figure}. This layout provides an intuitive way for users to refine system parameters while visualizing their effects. Key features of the PyQt application include:

\begin{figure}[h]
  \centering
  \includegraphics[width=.7\textwidth]{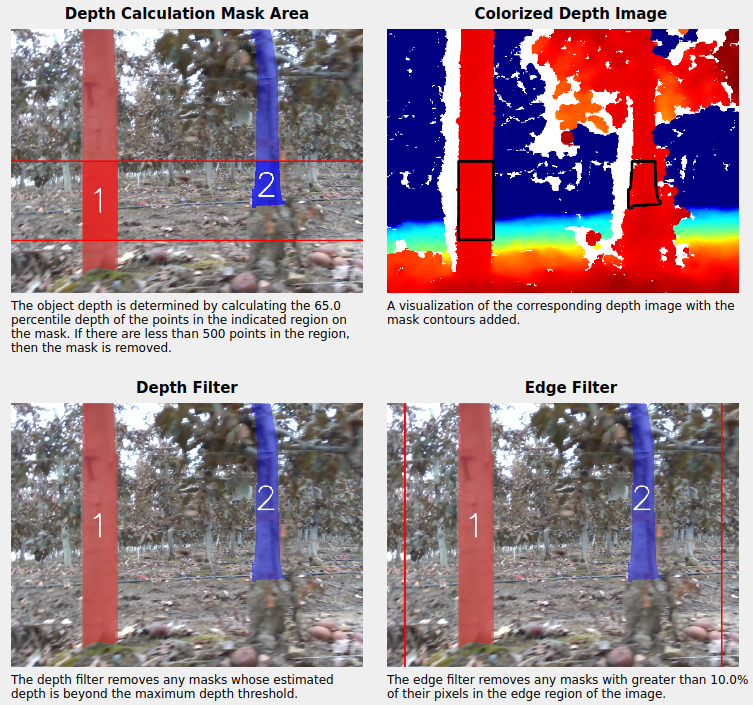}
  \caption{The tuning application displays visualizations of the trunk width estimation process for the current image, including explanations of each step. A subset of these visualization is shown here.}
  \label{fig:operation_visualization}
\end{figure}

\begin{itemize}
    \item \textbf{Parameter Tuning:} Users can select any operation in the pipeline and adjust its parameters interactively. The application immediately updates the displayed image to reflect the effect of the new parameters, allowing for immediate visual feedback.
    
    \item \textbf{Operation Toggling:} Some operations in the trunk width estimation process may be unnecessary for a given dataset and can be disabled to reduce processing time or assess their impact on the results. For example, users can toggle off filters that remove segmentations near the image edge, segmentations that are too high in the frame, or segmentations that are too far from the camera. Additionally, to reduce visual clutter, users can toggle individual operation visualizations displayed on the left side of the app.

    \item \textbf{Dataset Filtering:} The app allows users to filter the dataset based on specific features or parameter conditions. For example, users can filter to only include images containing certain objects (like a post), or to only include images where a particular filter operation removed a detection (to better understand the effect of that filter). Dataset filters can also isolate images based on parameters, e.g. those with detected trunks in specific regions of the image or with specific estimated depths.

    \item \textbf{Parameter Visualization:} To better understand the effects of various parameter changes, the app includes a histogram plotting feature. Users can visualize the distribution of key metrics such as estimated widths, detected trunk locations, or trunk depths across the dataset. This visualization aids in identifying potential outliers and refining parameter ranges.

    \item \textbf{Dataset Analysis:} After refining parameters, users can apply those parameters to the entire dataset. The app runs the full trunk width estimation pipeline on each image, reporting the resulting average error in width estimates across the dataset. This feature allows users to quantitatively assess how parameter changes impact overall system accuracy.
\end{itemize}

A sample dataset is provided online for convenience, with download instructions available on the project's GitHub repository. This dataset is also automatically included in the dev-container setup, simplifying the process for new users. These enhancements significantly improve the flexibility and usability of the trunk width estimation package, providing both developers and end-users with the tools needed to adapt the system to new orchard environments and datasets.

\subsection{Localization Module: Initial Prototype Design}

To simplify adoption of our orchard localization system, we developed a prototype (see Fig.~\ref{fig:prototype_module_diagram}) that integrates the key hardware and software components needed for deployment. This prototype was intended as a proof-of-concept for a platform that can be mounted on any orchard vehicle, enabling the localization system to function independently of vehicle type. The prototype incorporated visual odometry to reduce reliance on wheel odometry and included a remote display to improve usability. The system included the following primary components:

\begin{figure}[h]
  \centering
  \includegraphics[width=0.75\textwidth]{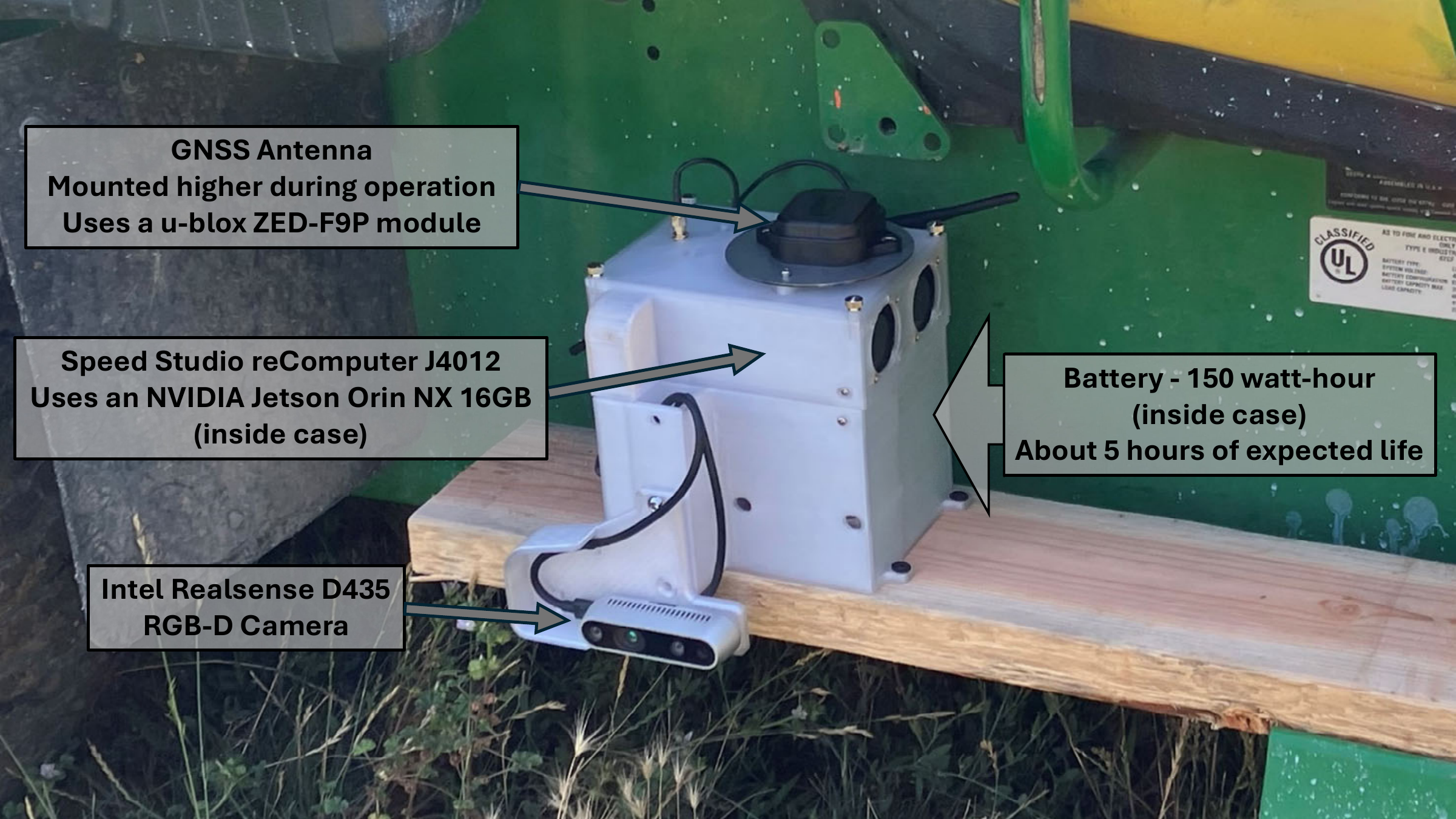}
  \caption{The prototype localization module mounted on an orchard utility vehicle. The prototype contains all of the sensing and computational hardware needed to perform localization, and has the ability to communicate with a remote device to display the location of the module on the map.}
  \label{fig:prototype_module_diagram}
\end{figure}

\begin{itemize}
    \item \textbf{Computing platform:} All computing was performed on a Seeed Studio (Shenzhen, China) reComputer J4012, single-board computer equipped with an NVIDIA (Santa Clara, CA, USA) Jetson Orin NX 16GB.
        
    \item \textbf{RGB-D camera for visual odometry:} The prototype used the RealSense D435 camera (Intel, Santa Clara, CA, USA) to capture RGB-D data; the camera was connected to the Orin and communicated with the localization system via ROS 2. We implemented visual odometry using the RGB-D data to provide linear displacement measurements. This method worked well for tracking movement in relatively straight lines and could accommodate gentle curves due to noise injected into the particle motion model. However, it struggled to track larger turns accurately (see Sec.~\ref{sec:Localization} for additional details).
 
    \item \textbf{GNSS sensor integration:} A SparkFun (Boulder, CO, USA) board with a ZED-F9P GNSS sensor was installed to provide additional localization data, though at this stage it was only used to approximate ground truth. This sensor was also connected to the Orin and communicated using ROS 2. The GNSS antenna has a magnetic base and a 10-foot cable, allowing it to be mounted directly on the prototype device or placed elsewhere on the vehicle for improved reception.
        
    \item \textbf{Remote display and control interface:} A Samsung (Suwon-si, South Korea) Galaxy Tab S9 FE was integrated as a remote display for user interaction and system control. The tablet offered a low-cost, water-resistant, and sunlight-readable solution for field use. The Orin display was streamed to the tablet using Virtual Network Computing (VNC), allowing the user to view and control the PyQt application running on the Orin.
        
    \item \textbf{Power supply:} The entire system was powered by a 12V LiFePO4 battery, which supported approximately five hours of field testing.
\end{itemize}

Depending on the exact configuration, the cost to procure and assemble the system is approximately \$1,850 USD. The prototype module was mounted on a utility vehicle for live localization testing. A video showcasing the module in action is available here: \url{https://www.youtube.com/watch?v=VH3EroQQkak}.
These tests demonstrated the system’s ability to perform accurate in-row localization during real-time navigation in a commercial orchard environment. However, the system struggled with tracking out-of-row turns accurately.

\subsection{Test Platform Integration and System Modifications for Evaluations}
\label{sec:Data Collection Setup}

\begin{figure}[h]
  \centering
  \includegraphics[width=0.75\textwidth]{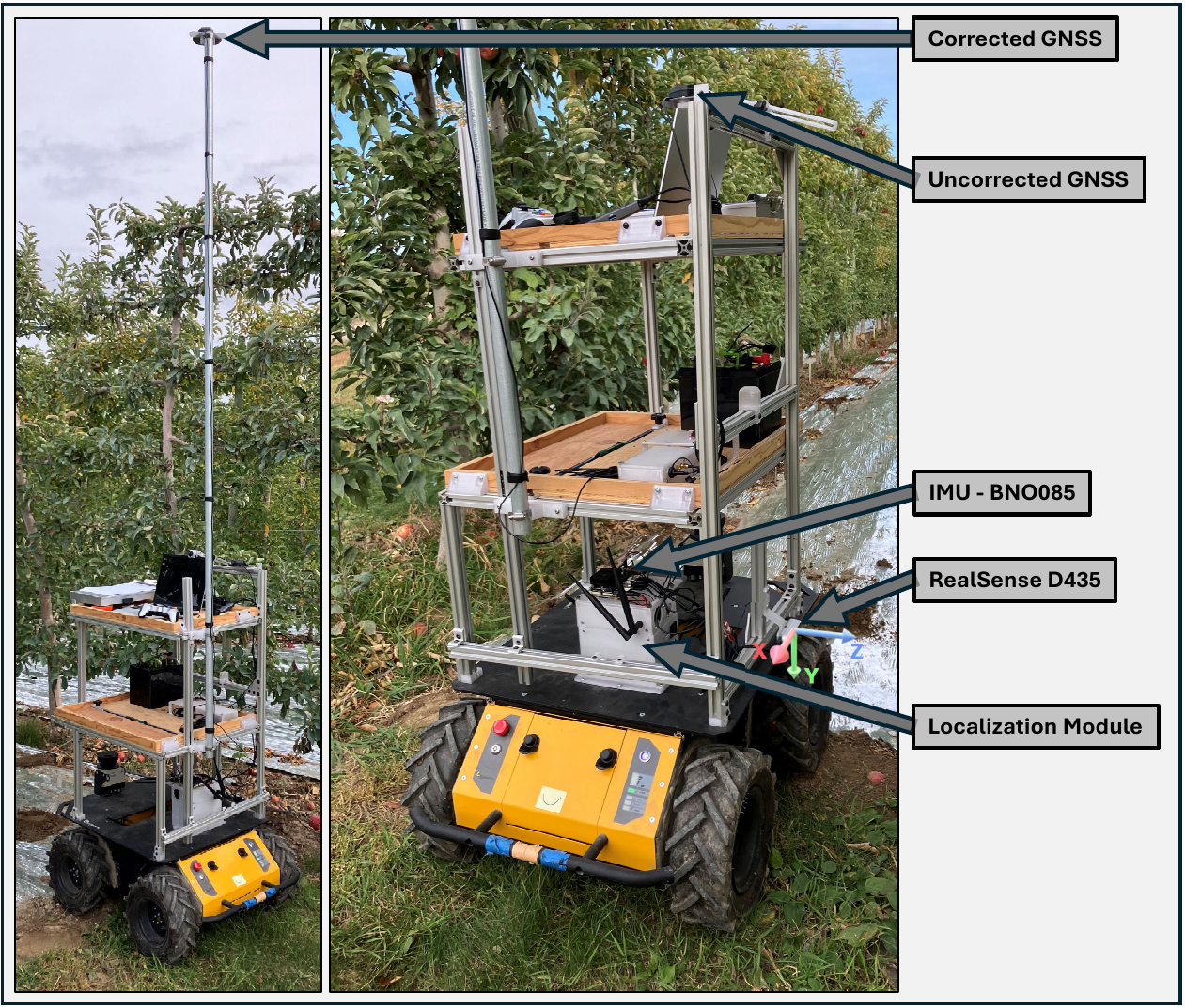}
  \caption{The setup used for collecting validation data with the Clearpath Husky. The camera axes are shown for reference. The camera's x-axis is parallel to the direction of forward/backward movement.}
  \label{fig:husky_setup}
\end{figure}

Following the initial prototype testing, we retained the standalone module but incorporated it into a comprehensive testing platform, making the following adjustments to i) improve localization accuracy during out-of-row headland turning; and ii) support quantitative evaluations of alternative motion models incorporating additional sensor inputs:

\begin{itemize}
    \item \textbf{Mounted system on a mobile robot:} The prototype module was mounted on a Clearpath Husky (Kitchener, ON, Canada) ground robot to serve as a stable testing platform. The Husky provided built-in wheel odometry and mounting space for sensors (see Fig.~\ref{fig:husky_setup}).
    
    \item \textbf{Integrated a BNO085 9-DOF IMU for orientation sensing:} An Adafruit (New York, NY, USA) BNO085 IMU breakout board was connected to the Orin and published orientation data via ROS 2. The IMU produced an orientation output by fusing magnetometer and gyroscope measurements. While the IMU also provides acceleration data, these measurements were not used.
    
    \item \textbf{Added an additional ZED-F9P GNSS:} To provide a ground truth reference for system evaluation, we added a second GNSS receiver configured to receive NTRIP corrections and mounted it on a 3.6 m mast for improved satellite visibility (see Fig.~\ref{fig:husky_setup}). We refer to this elevated, corrected GNSS as the `corrected' GNSS and the original, lower-mounted receiver without corrections as the `uncorrected' GNSS. The uncorrected GNSS, mounted at a height of 1.4 m, was retained as a basic positioning sensor and as a point of comparison against the corrected GNSS (see Sec.~\ref{sec:GNSS_evaluation}). Although the corrected GNSS could have been used as the system’s GNSS input, we opted to evaluate whether reliable localization was achievable using only the simpler configuration; providing NTRIP corrections and elevating an antenna are relatively straightforward tasks but still add complexity to a system. NTRIP corrections also require internet connectivity, which is not always available in remote locations.

\end{itemize}

In the following section, we describe several enhancements to our localization algorithm, followed by a description of the field experiments and evaluation metrics used to assess system performance.

\subsection{Updated Localization Algorithm}
\label{sec:Localization}

\textit{Preliminaries:} Our localization system implements a particle filter for online state estimation. Briefly, given a map of the environment, particle filter localization~(\cite{Dellaert_1999}) represents the vehicle’s possible poses with a set of weighted particles. The system recursively updates the particle states using a motion model and adjusts their weights based on sensor measurements. For a detailed description of our particle filter algorithm, the reader is referred to our original paper~(\cite{Brown_COMPAG_2024}); here we only describe new features added to improve the system's performance. 

To use SeeTree, a user requires a detailed orchard map of tree locations with associated trunk widths. For the present study, we used the map described in \cite{Brown_COMPAG_2024}. To account for growth of the trunks, all tree widths were uniformly increased by a small amount using a technique similar to that described in Section 3.1.3 of our original paper~(\cite{Brown_COMPAG_2024}). A map of the orchard can be created in several ways. For example, GNSS could be used to record tree and post locations when planting a new block when the trees are still very small. In an already established block with mature trees, SLAM-based techniques could be used to create the map. Additionally, while we did not originally design SeeTree for map making, it could be used for such purposes with minimal modifications. For example, it would be straightforward to populate a map with object locations using the trunk width estimation package, combined with the coordinate transformation to an elevated GNSS with NTRIP corrections as shown in Fig.~\ref{fig:husky_setup}.

\subsubsection{Motion Model}
\label{subsec:MotionModel}
To eliminate dependence on wheel encoders -- since many existing farm vehicles lack them -- we implemented two alternative motion models: one using the IMU for rotation and the camera for translation (visual odometry), and another using the IMU for rotation and GNSS for translation (GNSS odometry). Each method has distinct trade-offs. GNSS odometry requires a GNSS receiver but is computationally lightweight, making it effective in areas with reliable signal access. Although signal dropouts can occur under canopy or heavy cloud cover, we did not observe such issues during testing. Visual odometry, by contrast, uses only the onboard camera but has a higher computational cost.

\textbf{Particle rotation: IMU:} In the motion model, both newly added odometry methods (i.e., visual and GNSS) predict the rotation of the particles by adding Gaussian noise to the angular displacement, calculated as the difference between the IMU's orientation measurements at consecutive timestamps. 

\begin{figure}[h]
  \centering
  \includegraphics[width=0.95\textwidth]{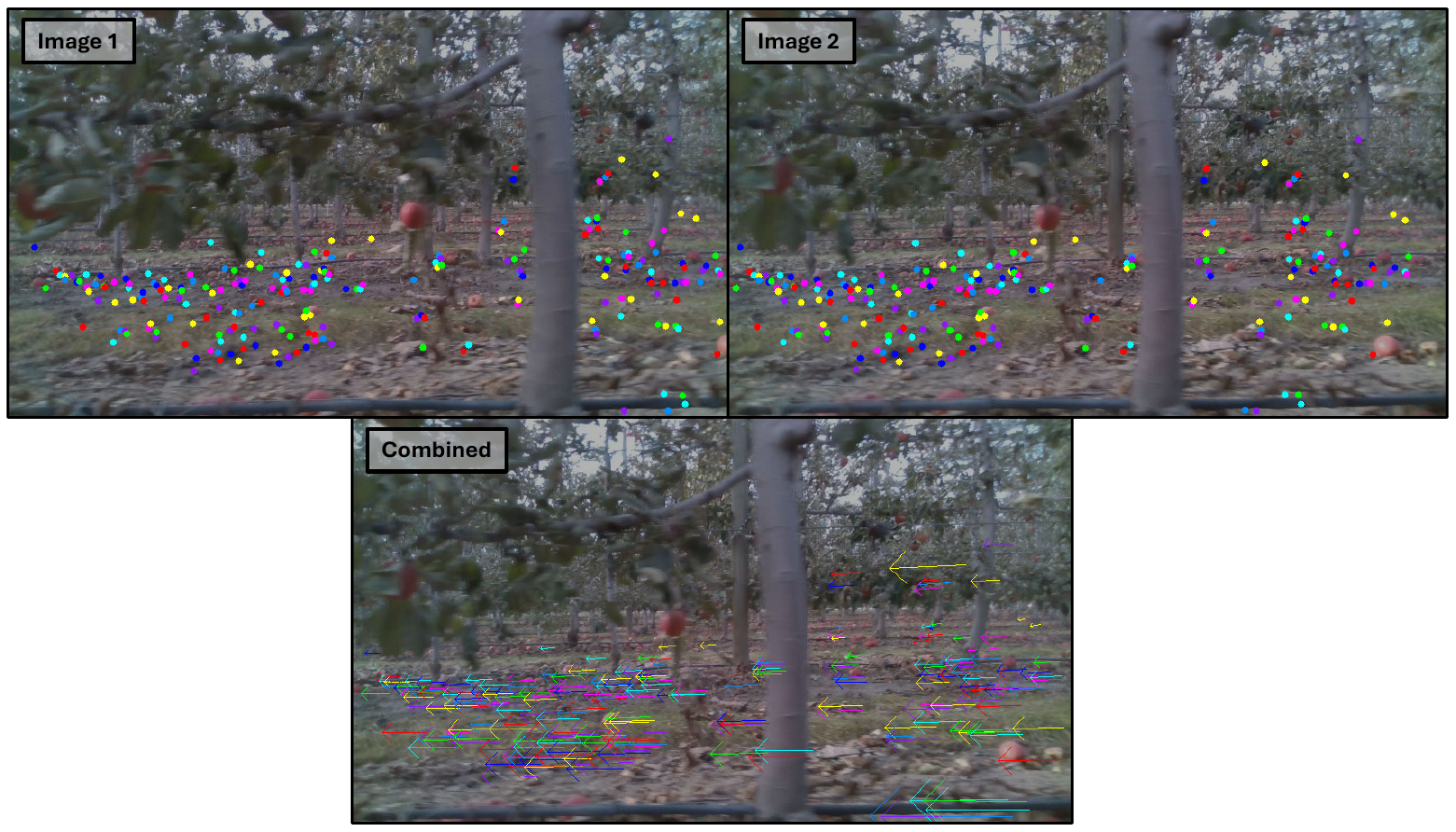}
  \caption{Translation estimation process for the visual odometry. Keypoints (colored dots) are detected and matched between two consecutive images, projected into 3D using depth data, and used to estimate camera movement. Only the component of displacement parallel to the camera’s x-axis is retained as the translation input for the motion model; rotational motion is handled separately using IMU orientation data.}
  \label{fig:visual_odometry}
\end{figure}

\textbf{Particle translation: Visual odometry:} There has been extensive prior work on visual odometry~(\cite{Scaramuzza_VOdometry_2011,Wang_VOdometry_2022}). Our visual odometry system uses the OpenCV implementation of the Scale-Invariant Feature Transform (SIFT) algorithm~(\cite{lowe2004distinctive}) to detect keypoints in each image from the RealSense D435 camera feed. These keypoints were matched to those found in the previous image (see Fig.~\ref{fig:visual_odometry}), and their corresponding 3D coordinates were calculated using the depth data. Camera movement was then estimated using OpenCV’s solvePnPRansac function, which computes a 3D translation and rotation vector from the matched points and camera parameters~(\cite{lepetit2009ep}). To simplify the motion estimate, only the component of the translation vector corresponding to forward/backward motion (parallel to the camera's x-axis as shown in Figure~\ref{fig:husky_setup}) was retained for use as the translation component of the visual odometry. This simplification was used because most movement occurred in this direction, while other motion components were either captured by the IMU's orientation measurements or accounted for by noise injected into the particle filter. 

\begin{figure}[h]
  \centering
  \includegraphics[width=0.5\textwidth]{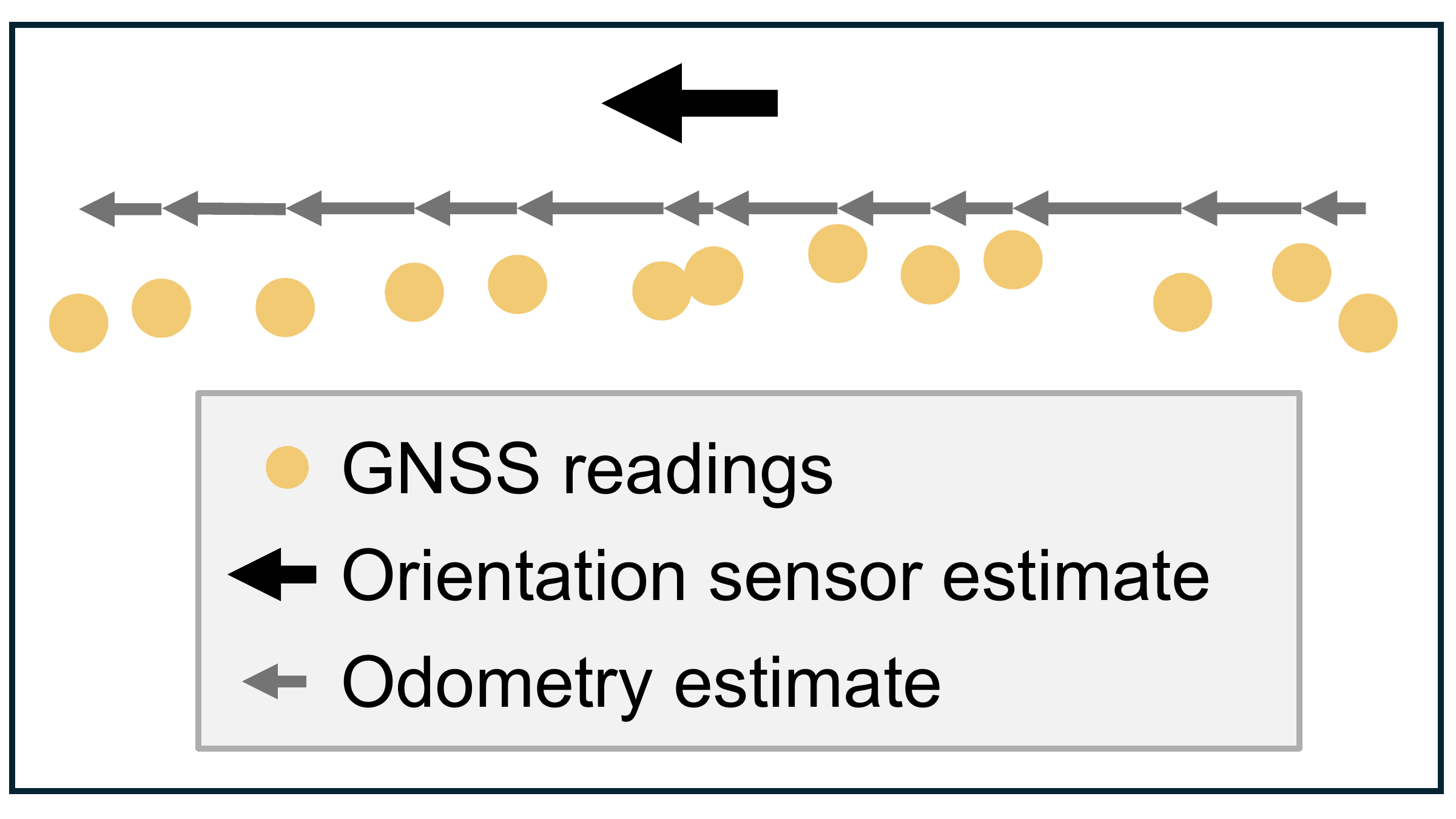}
  \caption{The GNSS odometry method estimates the translation distance by projecting a line between readings parallel to the output from the orientation sensor.}
  \label{fig:gnss_odom_explanation}
\end{figure}

\textbf{Particle translation: GNSS odometry:} \cite{Congram_GPS_2022} showed that accurate \textit{relative} position estimates can be found with one single-frequency GPS receiver. Our GNSS odometry system determines the total displacement between successive GNSS readings and keeps only the component parallel to the estimated vehicle heading, as determined by the IMU orientation sensor; see Fig.~\ref{fig:gnss_odom_explanation} for an illustration of the procedure. This method was used to reduce the effect of jitter in the GNSS readings caused by sensor error and vibrations in the Husky platform over rough terrain.

\subsubsection{Orientation Particle Weighting}
\label{subsec:ParticleWeighting}
In addition to being used to find the angular displacement in the odometry motion model, the IMU orientation sensor was also employed to weight the particles in the particle filter. The orientation weighting was implemented similarly to the trunk location and width weighting described in our previous work~(\cite{Brown_COMPAG_2024}), where sensor observations update the particle's likelihood. For orientation, the weight is computed using a Gaussian probability density function where the mean corresponds to the predicted orientation, and the standard deviation, set to 0.4 radians, represents the expected uncertainty in the orientation estimate. The actual observed orientation serves as the random variable, and the likelihood is determined by how closely the predicted orientation matches the observed one. This method and the chosen standard deviation are relatively lenient, preventing the removal of particles due to minor errors but removing those with large offsets.

\subsubsection{Why Both Angular Displacement and Orientation Are Useful}
\label{subsec:why_both}

In our system, angular displacement -- calculated as the change in the IMU’s orientation measurements between time steps -- is used in the motion model to update particle positions, while the orientation measurements themselves are used in the particle weighting step. While this might seem redundant, these two uses serve distinct roles in separate parts of the localization algorithm. The motion model determines particle movement and could, in future implementations, be combined with other motion data sources such as wheel encoders. The orientation output is used in the particle weighting step as one of several factors for evaluating particle likelihood. This weighting step also incorporates observations like tree range, bearing, and width -- and could be extended to include additional inputs. Although both rely on the same IMU data, their roles remain independent because each is treated differently with respect to noise: in the motion model, Gaussian noise is added to the calculated angular displacement to account for uncertainty in motion; in the particle weighting step, a separate Gaussian distribution defines the expected uncertainty in orientation, which determines how much an observed orientation error influences particle weights. This distinction ensures that the two uses are not redundant but instead provide complementary information that improves overall localization performance.

\section{Validation method}
\label{sec:Validation method}

To evaluate the performance of the improved localization system, we collected data using the setup shown in Fig.~\ref{fig:husky_setup}. We tested four odometry configurations: (i) the Husky's original wheel odometry (referred to as `plain' wheel odometry), (ii) wheel odometry with its angular displacement replaced by orientation estimates from the IMU (`IMU-augmented' wheel odometry), (iii) GNSS odometry, and (iv) visual odometry. In all configurations except plain wheel odometry, the same IMU orientation data was used to estimate angular motion.
Data from the IMU, GNSS sensors, RealSense camera, and Husky wheel odometry were recorded in ROS 2 bag files for analysis. Data was collected at the same commercial orchard plot used in our previous study~(\cite{Brown_COMPAG_2024}). 

\subsection{Data Collection Procedure}
\textbf{Full row traversals:} A total of 12 full-row runs were conducted. During these runs, the Husky started at one end of a row and was driven to the opposite end at a relatively constant speed (0.4 m/s) and orientation. These tests provided data for evaluating the system's ability to localize within a row when given a large initial pose uncertainty.

\textbf{Row changes:} In addition to full-row traversals, we conducted 43 row change turns. These involved the Husky starting near the end of one row, exiting that row, and re-entering either the same row or an adjacent row. These trials provided data for evaluating the system's ability to track turns and re-establish localization after a temporary loss of visual cues.

\subsection{Localization System Evaluation}

Using the collected data, we evaluated the system’s performance for three contexts: (i) in-row localization with a large initial uncertainty (900 m²), (ii) in-row localization with a small initial uncertainty (100 m²), and, (iii) maintaining localization during out-of-row turns. The evaluation procedures were adapted from the tuning tests described in Section 3.3 of our original paper~(\cite{Brown_COMPAG_2024}), which involved initializing particles in a defined area, propagating them forward using recorded data, and determining success based on whether the estimated position once the system converged was within an acceptable distance of the true location.

\begin{figure}[h]
  \centering
  \includegraphics[width=1.0\textwidth]{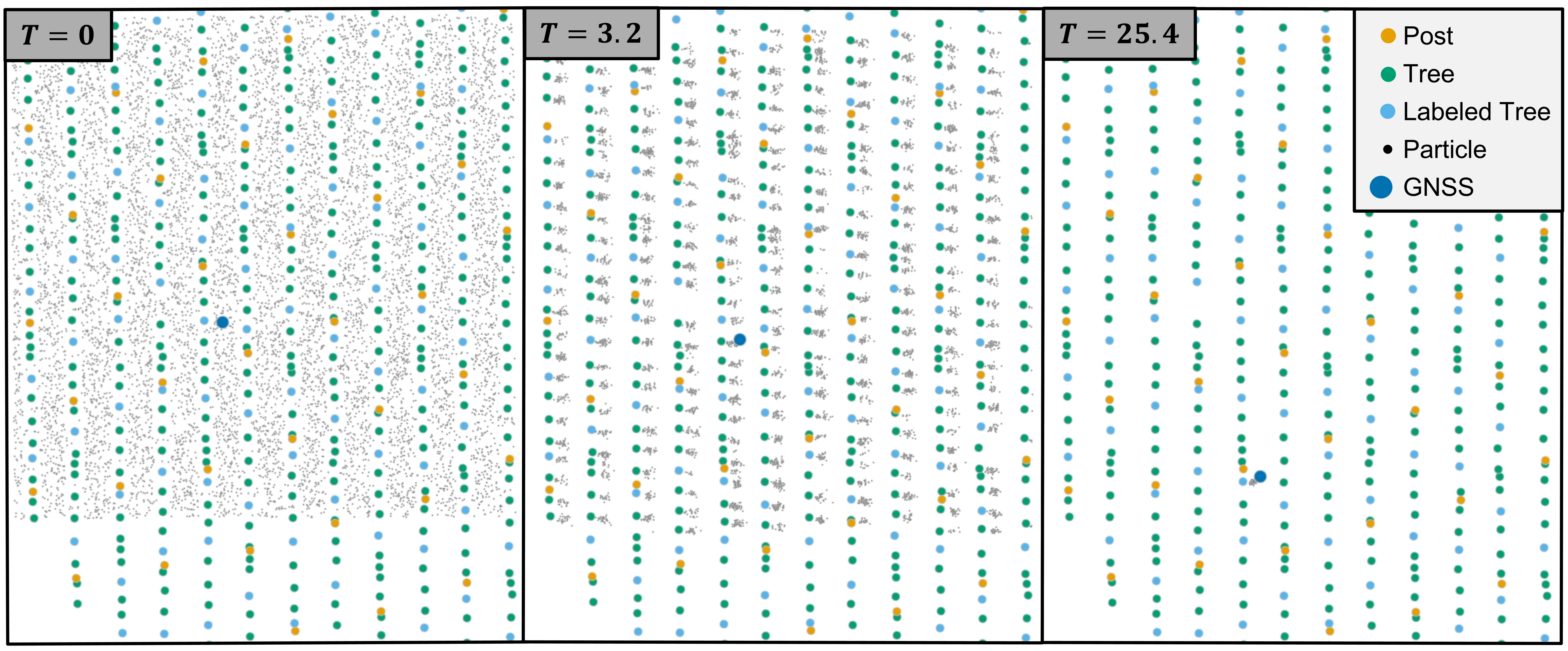}
  \caption{Progression of the particle filter during an evaluation trial within a row. The large blue dot represents the ground-truth GNSS reading, indicating the robot's actual position. At $T = 0\ \mathrm{s}$, particles are initialized within a 30 m x 30 m area. By $T = 3.2\ \mathrm{s}$, the system has observed a single tree, and by $T = 25.4\ \mathrm{s}$, the robot has moved far enough for the particle filter to converge on an accurate position estimate.}
  \label{fig:pf_progression_straight}
\end{figure}

\textbf{Evaluation along rows:} 
From the 12 full-row runs, 40 semi-random start locations were selected that provided a well-distributed sample set for testing. For each selected start location, 20 trials were conducted, resulting in a total of 800 trials for each odometry method. For each trial, the particles were initialized within a 900 m² area which encompassed the chosen starting point, with their orientations randomly set within ±5° of the row direction. Note, this initialization assumes that the Husky is generally facing down the row; while particles with larger incorrect orientations (i.e. greater than 5 degrees of error) would be quickly removed by the orientation sensor weighting, starting with mostly well-aligned particles avoids excess computation in the first few iterations. With the particles initialized, the recorded data was then used to propagate the particles forward. Once only one particle group remained, indicating convergence, the estimated position was taken as the highest-weighted particle. If the position was within 0.5 m of the actual location (see Sec.~\ref{sec:GNSS_evaluation}), the trial was considered a correct convergence. The total distance traveled between the starting and ending point was recorded as the distance-traveled metric for that trial. Figure~\ref{fig:pf_progression_straight} shows the progression of the particle filter during a typical trial. To simulate using an uncorrected GNSS for initializing the particle set, a second set of tests was completed where the initial starting area was reduced to 100 m² around the first GNSS reading.

\begin{figure}[h]
  \centering
  \includegraphics[width=0.8\textwidth]{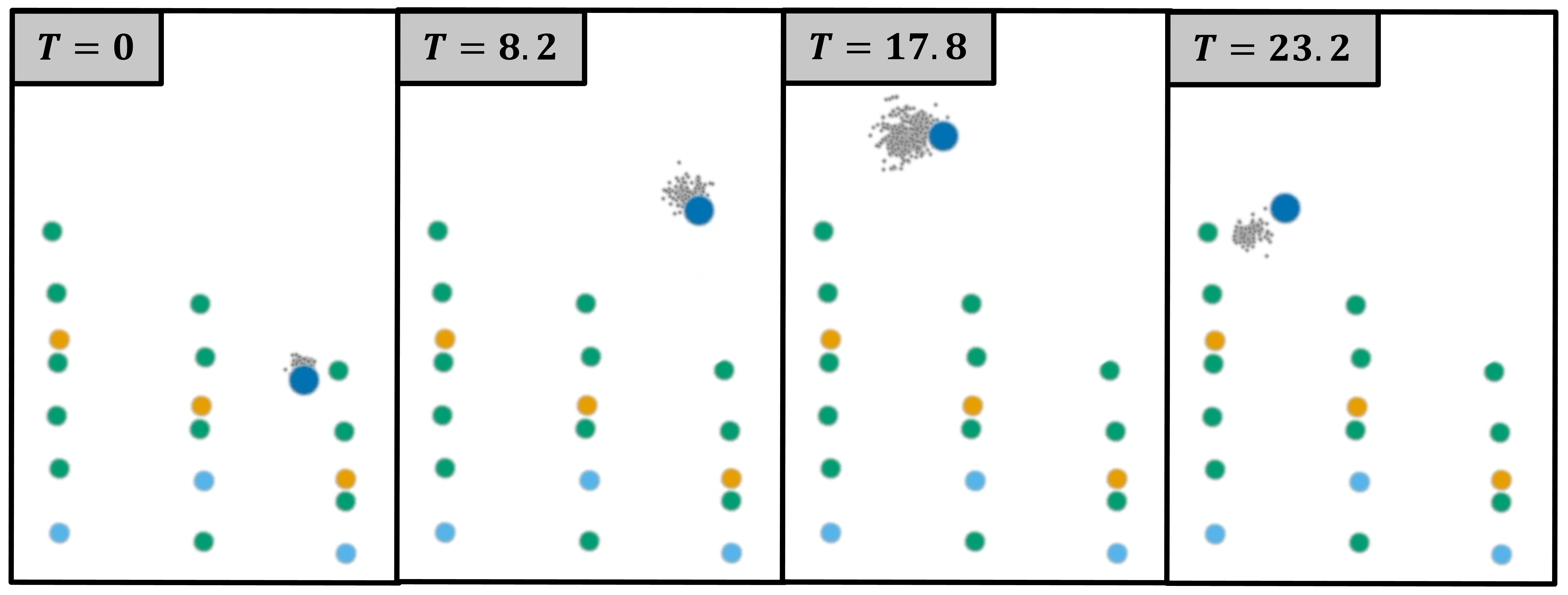}
  \caption{Progression of the particle filter during a turn. The large blue dot represents the GNSS reading, indicating the robot's actual position. At $T = 0\ \mathrm{s}$, the robot is preparing to exit the row. As it moves to the next row, the particle set spreads out due to the absence of visible trees. By $T = 23.2\ \mathrm{s}$, the robot has entered the next row, and the particle set has contracted based on the observation of a tree.}
  \label{fig:pf_progression_turn}
\end{figure}

\textbf{Evaluation for turns:} The second comparison examined the performance of each odometry source when tracking turns. For this evaluation, we used data from the 43 row changes. For each of these turns, 20 trials were conducted, resulting in a total of 860 trials for each analysis. Particles were initialized in a tight cluster centered around the Husky's known starting position. The recorded data was then used to propagate the particles forward through the turn. The trial was considered successful if the position estimate at the end of the turn accurately matched the Husky's actual position. Figure~\ref{fig:pf_progression_turn} shows the progression of the particle filter during an example turn.

\subsection{GNSS Evaluation}
\label{sec:GNSS_evaluation}
\begin{figure}[h]
  \centering
  \includegraphics[width=0.9\textwidth]{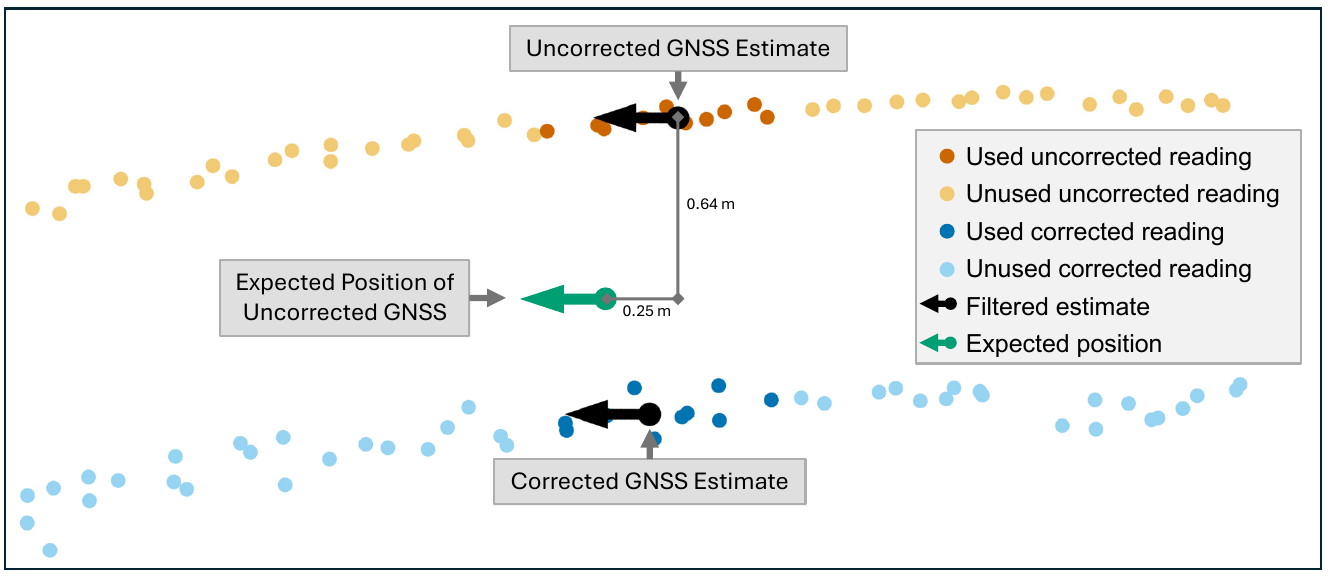}
  \caption{Visualization of GNSS sensor readings and corresponding filtered estimates. GNSS readings for a short travel distance are shown for both the uncorrected (orange) and corrected (blue) sensors. Darker points indicate readings used by the filter to generate the current estimates, marked by black arrows. The green arrow marks the expected position of the uncorrected GNSS sensor, based on the corrected GNSS reading and the known offset between the two sensors. At this time step, the uncorrected GNSS reading is offset by 0.25 m axially behind and 0.64 m transversely to the right of the expected position.}
  \label{fig:gnss_offset_explanation}
\end{figure}

To evaluate the performance of the uncorrected GNSS sensor in an orchard environment and assess its potential as a motion model source, we compared its output to that of the corrected GNSS, using the latter as ground truth. Our primary focus was on its drift rate—how quickly the error between the two sensors changed over time. This metric was chosen because, for motion estimation purposes, the relative accuracy between consecutive readings is more important than absolute positional accuracy. As described in Sec.~\ref{sec:System architecture}, both configurations used identical ZED-F9P sensors, but differed in antennae elevation and whether NTRIP corrections were applied. The uncorrected sensor was mounted approximately 1.0 m above the Clearpath Husky’s base, while the corrected sensor was mounted on a 3.2 m mast and received NTRIP corrections. Figure~\ref{fig:husky_setup} shows the sensor placement. While this evaluation was not intended as a comprehensive study, it provided a practical assessment of GNSS performance in a typical orchard setting. The following procedure was used to evaluate the uncorrected GNSS (see Fig.~\ref{fig:gnss_offset_explanation} for an illustration):

\begin{itemize}
    \item The latitude and longitude readings from each sensor were first converted to $X/Y$ coordinates in the orchard map’s reference frame.
    \item The resulting coordinates were smoothed using a degree-1 fitted spline 
    
    (\texttt{scipy.UnivariateSpline}, with $k=1$) applied across the 10 nearest time-stamped readings (5 before and 5 after each point). The orientation sensor output was used to determine the vehicle heading during offset calculations.
    \item For each data point, we calculated the expected position of the uncorrected GNSS based on the corrected GNSS reading, accounting for both sensors’ physical offsets. The difference between the expected and actual uncorrected GNSS position was measured along the transverse and axial directions relative to the Husky’s orientation. The total Euclidean distance was also recorded as an overall error metric.
    \item This process was repeated for all readings during the full-row runs. Additionally, the rate of change of the error was calculated to provide a measure of GNSS drift over time.
\end{itemize}

\section{Results and Discussion}
\label{sec:Results}

System performance was evaluated using the data and procedures described in Sec.~\ref{sec:Validation method}. These evaluations focused on three key aspects: i) how quickly and accurately the system converged to the correct position during straight row traversal; ii) its ability to maintain accurate localization through row-to-row turns; and iii) the suitability of uncorrected GNSS data as a source of short-term motion estimates. For the first two tasks, each test was repeated using the four odometry configurations: plain wheel odometry, IMU-augmented wheel odometry, visual odometry, and GNSS odometry. 

\subsection{Straight Row -- Large Start Area}

For these evaluations, we define accuracy as the percentage of trials in which the system successfully converged to within 0.5 m of the robot’s actual position. As shown in Table~\ref{table:large-area_results}, visual and GNSS odometry both achieved nearly perfect accuracy, outperforming plain wheel odometry. IMU-augmented wheel odometry also performed well, with accuracy nearly matching that of visual and GNSS odometry.

In terms of convergence distance, visual and GNSS odometry resulted in average travel distances of 9.43 m and 9.46 m, respectively, slightly better than IMU-augmented wheel odometry (10.69 m) and plain wheel odometry (10.52 m). These results demonstrate that both visual and GNSS odometry are viable alternatives for localization in orchard environments. The standard deviations are relatively large because convergence distance depends heavily on starting location -- e.g., trials that began near a unique feature, such as two large trees close together, converged more quickly.

\begin{table}[H]
\caption{Accuracy, the average distance traveled before convergence, and associated standard deviation for the 800 trials for each odometry method when initialized in a 900 m² area.}
\vspace{-5pt}
\centering
\small
\begin{tabular}{lccc}
\hline
\textbf{Odometry} & \textbf{Accuracy} & \textbf{Distance (m)} & \textbf{Std. Dev. (m)} \\
\hline
Wheel & 0.925 & 10.52 & 3.55 \\
Wheel w/ IMU & 0.995 & 10.69 & 3.71 \\
Visual & 1.000 & 9.43 & 3.15 \\
GNSS & 0.996 & 9.46 & 3.12 \\
\hline
\label{table:large-area_results}
\end{tabular}
\end{table}

\subsection{Straight Row -- Small Start Area}

When particles were initialized in a smaller 100 m² area, accuracy rates remained consistent with the previous trials, as shown in Table~\ref{table:small-area_results}. However, the distance traveled before convergence was reduced across all methods. Visual odometry averaged 6.28 m, GNSS odometry averaged 6.20 m, IMU-augmented wheel odometry averaged 6.44 m, and plain wheel odometry averaged 6.52 m. This demonstrates that starting with a more confident pose estimate improves convergence speed.

\begin{table}[H]
\caption{Accuracy, the average distance traveled before convergence, and associated standard deviation for the 800 trials for each odometry method when initialized in a 100 m² area around the first GNSS reading.}
\vspace{-5pt} 
\centering
\small
\begin{tabular}{lccc}
\hline
\textbf{Odometry} & \textbf{Accuracy} & \textbf{Distance (m)} & \textbf{Std. Dev. (m)} \\
\hline
Wheel & 0.956 & 6.52 & 3.34 \\
Wheel w/ IMU & 0.999 & 6.44 & 2.95 \\
Visual & 0.996 & 6.28 & 2.79 \\
GNSS & 0.999 & 6.20 & 2.80 \\
\hline
\label{table:small-area_results}
\end{tabular}
\end{table}

\subsection{Row Changes}

For the row-to-row turn tests, a trial was considered successful if the estimated final position matched the robot’s true position at the end of the turn. As shown in Table~\ref{table:turn_results}, plain wheel odometry performed the worst, with an accuracy of 0.786. Visual odometry achieved 0.938, while IMU-augmented wheel odometry and GNSS odometry performed best, with accuracies of 0.974 and 0.999, respectively. These results suggest that visual, GNSS, and IMU-augmented wheel odometry each provided reasonably accurate tracking through turns, especially compared to plain wheel odometry.

\begin{table}[H]
\caption{Accuracy of system when tracking row-to-row turns using different odometry methods.}
\vspace{-5pt} 
\centering
\small
\begin{tabular}{lc}
\hline
\textbf{Odometry} & \textbf{Accuracy} \\
\hline
Wheel & 0.786 \\
Wheel w/ IMU & 0.974 \\
Visual & 0.938 \\
GNSS & 0.999 \\
\hline
\label{table:turn_results}
\end{tabular}
\end{table}

In summary, all three new odometry methods showed strong performance compared to the plain wheel odometry. They consistently outperformed it in both convergence distance and accuracy, demonstrating that visual and GNSS odometry are viable alternatives to traditional wheel odometry and may provide improved performance in certain conditions. Additionally, the improved performance of IMU-augmented wheel odometry demonstrates the value of integrating angular displacement data from an independent sensor.

While our visual odometry technique performed well, it is relatively basic and required hand-tuning for the orchard environment selected for this study. Much of the recent work in visual odometry incorporates deep learning~(\cite{Wang_VO_2017,Li_UnDeepVO_2018,Yang_2020_CVPR,Wang_VOdometry_2022}) to improve upon the brittleness of classical techniques and automatically learn effective representations from data. In the future, such visual odometry techniques could be integrated in SeeTree to improve localization robustness and overall system efficiency. 

Another opportunity for performance improvements is combining the odometry methods, such as integrating visual-inertial odometry~(\cite{Geneva_OpenVins_2020,Qin_VinsMono_2018,Campos_ORBSLAM3_2021}), for a more accurate motion model. More recently, \cite{Zhang_Filter_2025} presented an invariant filter that fuses Visual-Inertial-GNSS-Barometer odometry (GBVIO). The open-source and modular architecture of SeeTree makes the implementation of new algorithms and components relatively straightforward. 

\begin{figure}[h]
  \centering
  \includegraphics[width=0.8\textwidth]{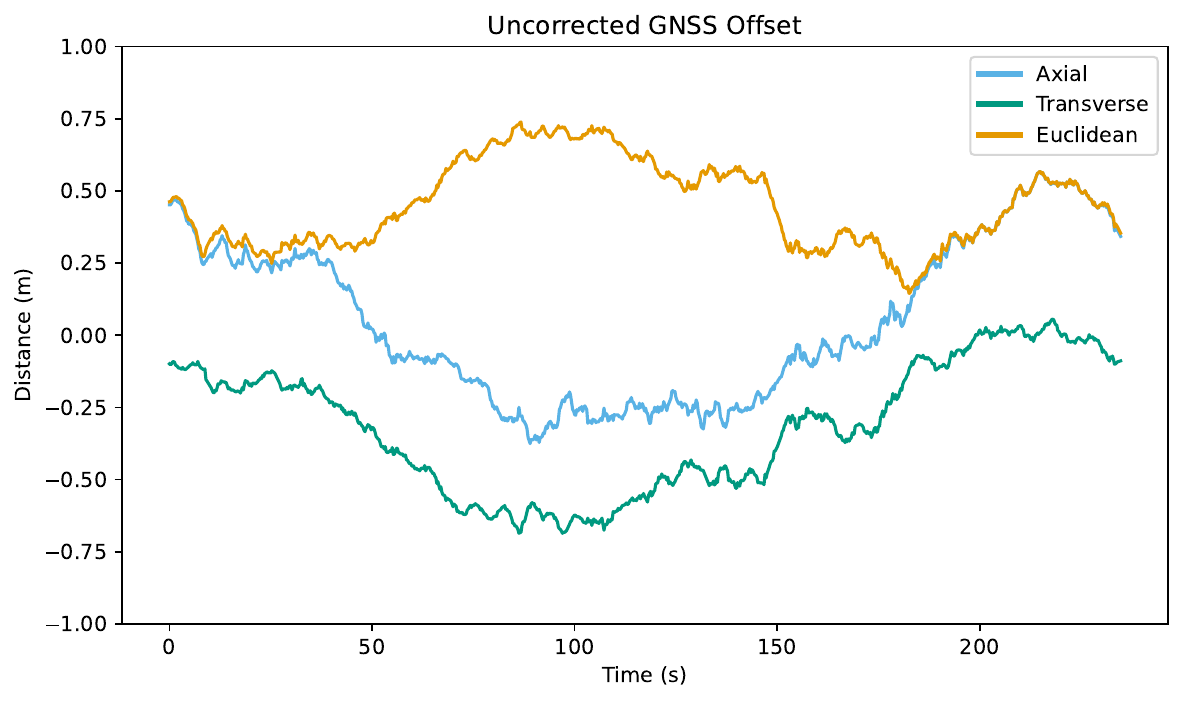}
  \caption{The axial, transverse, and total Euclidean offsets between the uncorrected and corrected GNSS readings during a single row traversal. The total duration was 234 seconds, with the platform generally moving at approximately 0.4 m/s.}
  \label{fig:offset_figure}
\end{figure}

\subsection{Corrected vs. Uncorrected GNSS Comparison}
\begin{figure}[h]
  \centering
  \includegraphics[width=0.8\textwidth]{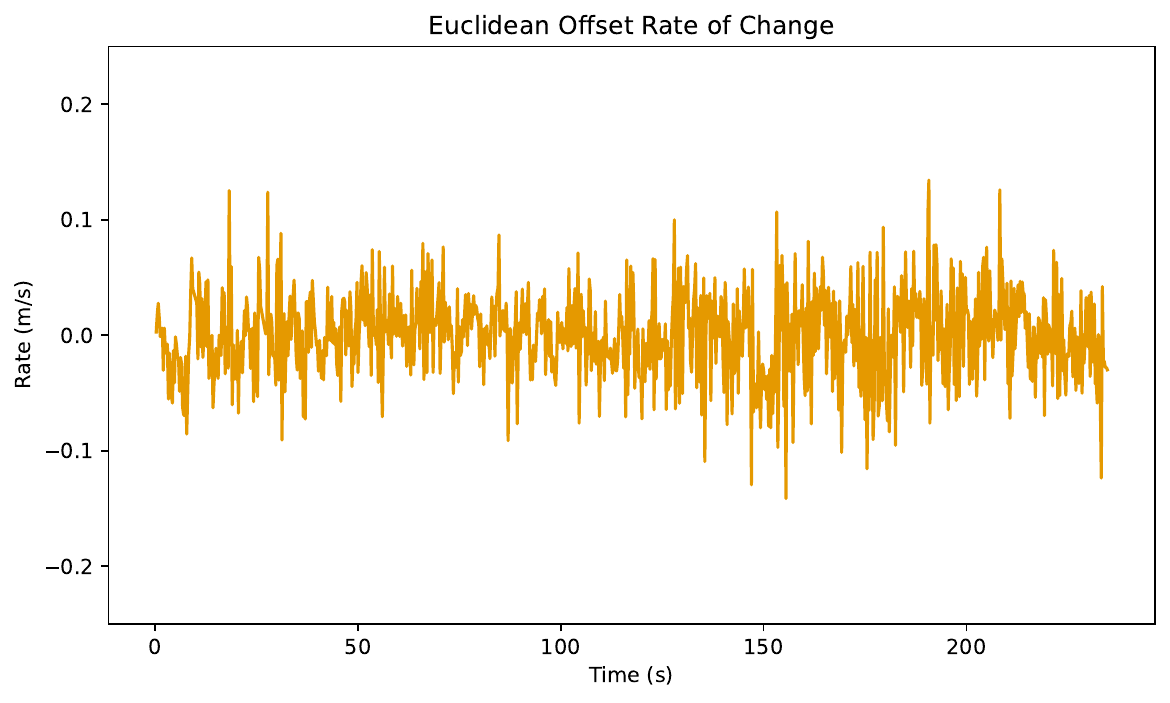}
  \caption{The rate of change of the total Euclidean offset shown in Figure \ref{fig:offset_figure}.}
  \label{fig:offset_delta_figure}
\end{figure}

\begin{figure}
  \centering
  \includegraphics[width=0.9\textwidth]{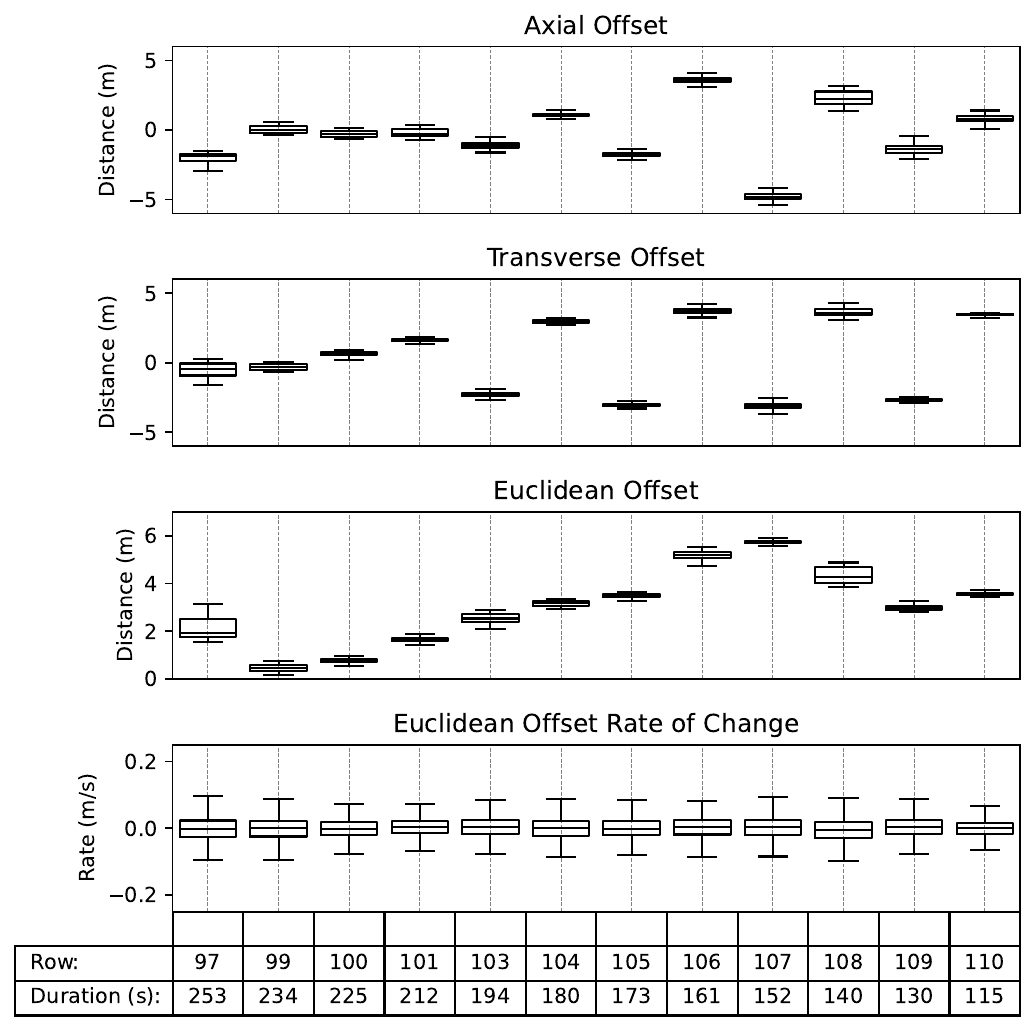}
  \caption{The top three plots show the distribution of axial, transverse, and total Euclidean offsets between the uncorrected and corrected GNSS measurements across all full row runs. The median in each box plot is the median offset for that run, and the range illustrates how much the offset varied. The bottom plot shows the rate of change of the total Euclidean offset for each run. The data presented in Figs.~\ref{fig:offset_figure} and \ref{fig:offset_delta_figure} is from Row 99.}
  \label{fig:offset_box_plot}
\end{figure}

Figures~\ref{fig:offset_figure}–\ref{fig:offset_box_plot} show results from the analysis of the uncorrected GNSS's drift rate and accuracy. Figure~\ref{fig:offset_figure} shows the axial, transverse, and total Euclidean offsets between the  uncorrected and corrected GNSS measurements during a single row traversal. The total run time was 234 seconds, with the platform generally moving at about 0.4 m/s. While the offset fluctuated throughout the run, these fluctuations were gradual, as shown in Figure \ref{fig:offset_delta_figure} which depicts the rate of change of the offset during the run. The sporadic nature of that plot is primarily due to antenna movement and oscillations. These oscillations are evident in Figure \ref{fig:gnss_offset_explanation}, where the corrected GNSS (blue dots) exhibits greater variability than the uncorrected GNSS (orange dots), likely due to the corrected GNSS being mounted on a tall mast that was susceptible to sway. Despite this movement, the overall magnitude of the offset's rate of change was low. Figure \ref{fig:offset_box_plot} presents this data condensed into box-and-whisker plots for all recorded row traversals (including the example shown in Figure \ref{fig:offset_figure}). The top three plots show the variance in axial, transverse, and total Euclidean offsets. The bottom plot illustrates the rate of change of the total Euclidean offset.


From these results, it is clear that while the magnitude of the uncorrected GNSS's error does change over time, the rate of change remains relatively low. This stability suggests that the uncorrected GNSS can serve as an effective odometry source. While the uncorrected GNSS may drift over time and is unsuitable for precise absolute positioning, the incremental movement between individual readings closely matches the robot's actual movement. As a result, the uncorrected GNSS is well-suited for providing short-term motion estimates within our localization system, even if its absolute position accuracy degrades over time.

We did not complete experiments/evaluations with the uncorrected GNSS for extended durations (i.e. days), so while the system performed well the entire time that it was tested, how well it works long term is unknown. Geomagnetic storms in space would most likely be a significant problem for our GNSS odometry technique. For such situations, it may be possible to implement logic that detects unrealistic shifts in GNSS measurements based on the vehicle's dynamics and then shifts to an alternate odometry source such as visual odometry.

\section{Conclusion}
\label{sec:Conclusion}

In this paper, we presented SeeTree, an open-source system for tree segmentation, automatic trunk width measurement, and accurate spatial localization that can be used with any vehicle, autonomous or human-driven. SeeTree uses off-the-shelf components, includes visualization tools that can be used to optimize system performance for new environments, and includes support for multiple types of odometry. Extensive field experiments showed that GNSS odometry and wheel odometry augmented with an imu were the most accurate odometry sources during in-row localization, and that GNSS odometry was the most accurate during out-of-row turning. The least accurate odometry was plain wheel odometry. The system with the highest performing configuration converged to the correct within-row location 99\% of the time, usually within 6 m if starting with a more confident initial pose estimate, and was also able to track 99\% of sequential row turns across 860 trials involving 43 unique turns. Future work will focus on fusing odometry methods for improved efficiency and robustness across orchard environments. To support adoption and future development, we make the source code and dataset freely available to the community.

\section{Acknowledgements}
This research was supported by the Washington Tree Fruit Research Commission. We also thank Allan Brothers Fruit Inc. (Naches, WA, USA) for allowing us to conduct experiments at their commercial orchards. 

\bibliographystyle{elsarticle-harv} 
\bibliography{orchard_localization}

\begin{thebibliography}{21}
\expandafter\ifx\csname natexlab\endcsname\relax\def\natexlab#1{#1}\fi
\providecommand{\url}[1]{\texttt{#1}}
\providecommand{\href}[2]{#2}
\providecommand{\path}[1]{#1}
\providecommand{\DOIprefix}{doi:}
\providecommand{\ArXivprefix}{arXiv:}
\providecommand{\URLprefix}{URL: }
\providecommand{\Pubmedprefix}{pmid:}
\providecommand{\doi}[1]{\href{http://dx.doi.org/#1}{\path{#1}}}
\providecommand{\Pubmed}[1]{\href{pmid:#1}{\path{#1}}}
\providecommand{\bibinfo}[2]{#2}
\ifx\xfnm\relax \def\xfnm[#1]{\unskip,\space#1}\fi
\bibitem[{Aguiar et~al.(2020)Aguiar, dos Santos, Cunha, Sobreira and Sousa}]{Aguiar_Robotics_2020}
\bibinfo{author}{Aguiar, A.S.}, \bibinfo{author}{dos Santos, F.N.}, \bibinfo{author}{Cunha, J.B.}, \bibinfo{author}{Sobreira, H.}, \bibinfo{author}{Sousa, A.J.}, \bibinfo{year}{2020}.
\newblock \bibinfo{title}{Localization and mapping for robots in agriculture and forestry: A survey}.
\newblock \bibinfo{journal}{Robotics} \bibinfo{volume}{9}.
\newblock \URLprefix \url{https://www.mdpi.com/2218-6581/9/4/97}, \DOIprefix\doi{10.3390/robotics9040097}.
\bibitem[{Brown et~al.(2024)Brown, Paudel, Biehler, Thompson, Karkee, Grimm and Davidson}]{Brown_COMPAG_2024}
\bibinfo{author}{Brown, J.}, \bibinfo{author}{Paudel, A.}, \bibinfo{author}{Biehler, D.}, \bibinfo{author}{Thompson, A.}, \bibinfo{author}{Karkee, M.}, \bibinfo{author}{Grimm, C.}, \bibinfo{author}{Davidson, J.R.}, \bibinfo{year}{2024}.
\newblock \bibinfo{title}{Tree detection and in-row localization for autonomous precision orchard management}.
\newblock \bibinfo{journal}{Computers and Electronics in Agriculture} \bibinfo{volume}{227}, \bibinfo{pages}{109454}.
\newblock \DOIprefix\doi{https://doi.org/10.1016/j.compag.2024.109454}.
\bibitem[{Campos et~al.(2021)Campos, Elvira, Rodríguez, M.~Montiel and D.~Tardós}]{Campos_ORBSLAM3_2021}
\bibinfo{author}{Campos, C.}, \bibinfo{author}{Elvira, R.}, \bibinfo{author}{Rodríguez, J.J.G.}, \bibinfo{author}{M.~Montiel, J.M.}, \bibinfo{author}{D.~Tardós, J.}, \bibinfo{year}{2021}.
\newblock \bibinfo{title}{Orb-slam3: An accurate open-source library for visual, visual–inertial, and multimap slam}.
\newblock \bibinfo{journal}{IEEE Transactions on Robotics} \bibinfo{volume}{37}, \bibinfo{pages}{1874--1890}.
\bibitem[{Congram and Barfoot(2022)}]{Congram_GPS_2022}
\bibinfo{author}{Congram, B.}, \bibinfo{author}{Barfoot, T.D.}, \bibinfo{year}{2022}.
\newblock \bibinfo{title}{Field testing and evaluation of single-receiver gps odometry for use in robotic navigation}.
\newblock \bibinfo{journal}{Field Robotics} \bibinfo{volume}{2}, \bibinfo{pages}{1849--1873}.
\bibitem[{Dellaert et~al.(1999)Dellaert, Fox, Burgard and Thrun}]{Dellaert_1999}
\bibinfo{author}{Dellaert, F.}, \bibinfo{author}{Fox, D.}, \bibinfo{author}{Burgard, W.}, \bibinfo{author}{Thrun, S.}, \bibinfo{year}{1999}.
\newblock \bibinfo{title}{Monte carlo localization for mobile robots}, in: \bibinfo{booktitle}{Proceedings 1999 IEEE International Conference on Robotics and Automation (Cat. No.99CH36288C)}, pp. \bibinfo{pages}{1322--1328 vol.2}.
\bibitem[{Gallardo et~al.(2019)Gallardo, Grant, Brown, McFerson, Lewis, Einhorn and Sazo}]{PerceptionsofPrecisionAgricultureTechnologiesintheUSFreshAppleIndustry}
\bibinfo{author}{Gallardo, R.K.}, \bibinfo{author}{Grant, K.}, \bibinfo{author}{Brown, D.J.}, \bibinfo{author}{McFerson, J.R.}, \bibinfo{author}{Lewis, K.M.}, \bibinfo{author}{Einhorn, T.}, \bibinfo{author}{Sazo, M.M.}, \bibinfo{year}{2019}.
\newblock \bibinfo{title}{Perceptions of precision agriculture technologies in the u.s. fresh apple industry}.
\newblock \bibinfo{journal}{HortTechnology} \bibinfo{volume}{29}, \bibinfo{pages}{151 -- 162}.
\bibitem[{Geneva et~al.(2020)Geneva, Eckenhoff, Lee, Yang and Huang}]{Geneva_OpenVins_2020}
\bibinfo{author}{Geneva, P.}, \bibinfo{author}{Eckenhoff, K.}, \bibinfo{author}{Lee, W.}, \bibinfo{author}{Yang, Y.}, \bibinfo{author}{Huang, G.}, \bibinfo{year}{2020}.
\newblock \bibinfo{title}{Openvins: A research platform for visual-inertial estimation}, in: \bibinfo{booktitle}{2020 IEEE International Conference on Robotics and Automation (ICRA)}, pp. \bibinfo{pages}{4666--4672}.
\bibitem[{Guevara et~al.(2024)Guevara, Gené-Mola, Gregorio and {Auat Cheein}}]{Guevara_SmartAgTech_2024}
\bibinfo{author}{Guevara, J.}, \bibinfo{author}{Gené-Mola, J.}, \bibinfo{author}{Gregorio, E.}, \bibinfo{author}{{Auat Cheein}, F.A.}, \bibinfo{year}{2024}.
\newblock \bibinfo{title}{A systematic analysis of scan matching techniques for localization in dense orchards}.
\newblock \bibinfo{journal}{Smart Agricultural Technology} \bibinfo{volume}{9}, \bibinfo{pages}{100607}.
\newblock \DOIprefix\doi{https://doi.org/10.1016/j.atech.2024.100607}.
\bibitem[{Lepetit et~al.(2009)Lepetit, Moreno-Noguer and Fua}]{lepetit2009ep}
\bibinfo{author}{Lepetit, V.}, \bibinfo{author}{Moreno-Noguer, F.}, \bibinfo{author}{Fua, P.}, \bibinfo{year}{2009}.
\newblock \bibinfo{title}{Ep n p: An accurate o (n) solution to the p n p problem}.
\newblock \bibinfo{journal}{International journal of computer vision} \bibinfo{volume}{81}, \bibinfo{pages}{155--166}.
\bibitem[{Li et~al.(2018)Li, Wang, Long and Gu}]{Li_UnDeepVO_2018}
\bibinfo{author}{Li, R.}, \bibinfo{author}{Wang, S.}, \bibinfo{author}{Long, Z.}, \bibinfo{author}{Gu, D.}, \bibinfo{year}{2018}.
\newblock \bibinfo{title}{Undeepvo: Monocular visual odometry through unsupervised deep learning}, in: \bibinfo{booktitle}{2018 IEEE International Conference on Robotics and Automation (ICRA)}, pp. \bibinfo{pages}{7286--7291}.
\bibitem[{Lowe(2004)}]{lowe2004distinctive}
\bibinfo{author}{Lowe, D.G.}, \bibinfo{year}{2004}.
\newblock \bibinfo{title}{Distinctive image features from scale-invariant keypoints}.
\newblock \bibinfo{journal}{International journal of computer vision} \bibinfo{volume}{60}, \bibinfo{pages}{91--110}.
\bibitem[{McFadden et~al.(February 2023)McFadden, Njuki and Griffin}]{PrecisionAg_2023}
\bibinfo{author}{McFadden, J.}, \bibinfo{author}{Njuki, E.}, \bibinfo{author}{Griffin, T.}, \bibinfo{year}{February 2023}.
\newblock \bibinfo{title}{Precision Agriculture in the Digital Era: Recent Adoption on U.S. Farms}.
\newblock \bibinfo{type}{Technical Report}. U.S. Department of Agriculture Economic Research Service.
\bibitem[{Qin et~al.(2018)Qin, Li and Shen}]{Qin_VinsMono_2018}
\bibinfo{author}{Qin, T.}, \bibinfo{author}{Li, P.}, \bibinfo{author}{Shen, S.}, \bibinfo{year}{2018}.
\newblock \bibinfo{title}{Vins-mono: A robust and versatile monocular visual-inertial state estimator}.
\newblock \bibinfo{journal}{IEEE Transactions on Robotics} \bibinfo{volume}{34}, \bibinfo{pages}{1004--1020}.
\bibitem[{Scaramuzza and Fraundorfer(2011)}]{Scaramuzza_VOdometry_2011}
\bibinfo{author}{Scaramuzza, D.}, \bibinfo{author}{Fraundorfer, F.}, \bibinfo{year}{2011}.
\newblock \bibinfo{title}{Visual odometry [tutorial]}.
\newblock \bibinfo{journal}{IEEE Robotics \& Automation Magazine} \bibinfo{volume}{18}, \bibinfo{pages}{80--92}.
\bibitem[{Sivakumar et~al.(2024)Sivakumar, Gasparino, McGuire, Higuti, Akcal and Chowdhary}]{Sivakumar-RSS-24}
\bibinfo{author}{Sivakumar, A.N.}, \bibinfo{author}{Gasparino, M.V.}, \bibinfo{author}{McGuire, M.}, \bibinfo{author}{Higuti, V.A.H.}, \bibinfo{author}{Akcal, M.U.}, \bibinfo{author}{Chowdhary, G.}, \bibinfo{year}{2024}.
\newblock \bibinfo{title}{{Demonstrating CropFollow++: Robust Under-Canopy Navigation with Keypoints}}, in: \bibinfo{booktitle}{Proceedings of Robotics: Science and Systems}, \bibinfo{address}{Delft, Netherlands}.
\newblock \DOIprefix\doi{10.15607/RSS.2024.XX.023}.
\bibitem[{Tang et~al.(2024)Tang, Guo, Huang, Huai and Gai}]{Tang_IEEE_2024}
\bibinfo{author}{Tang, B.}, \bibinfo{author}{Guo, Z.}, \bibinfo{author}{Huang, C.}, \bibinfo{author}{Huai, S.}, \bibinfo{author}{Gai, J.}, \bibinfo{year}{2024}.
\newblock \bibinfo{title}{A fruit-tree mapping system for semi-structured orchards based on multi-sensor-fusion slam}.
\newblock \bibinfo{journal}{IEEE Access} \bibinfo{volume}{12}, \bibinfo{pages}{162122--162130}.
\newblock \DOIprefix\doi{10.1109/ACCESS.2024.3408467}.
\bibitem[{Underwood et~al.(2015)Underwood, Jagbrant, Nieto and Sukkarieh}]{Underwood_JFR_2015}
\bibinfo{author}{Underwood, J.P.}, \bibinfo{author}{Jagbrant, G.}, \bibinfo{author}{Nieto, J.I.}, \bibinfo{author}{Sukkarieh, S.}, \bibinfo{year}{2015}.
\newblock \bibinfo{title}{Lidar-based tree recognition and platform localization in orchards}.
\newblock \bibinfo{journal}{Journal of Field Robotics} \bibinfo{volume}{32}, \bibinfo{pages}{1056--1074}.
\bibitem[{Wang et~al.(2022)Wang, Ma, Chen, Ren and Lu}]{Wang_VOdometry_2022}
\bibinfo{author}{Wang, K.}, \bibinfo{author}{Ma, S.}, \bibinfo{author}{Chen, J.}, \bibinfo{author}{Ren, F.}, \bibinfo{author}{Lu, J.}, \bibinfo{year}{2022}.
\newblock \bibinfo{title}{Approaches, challenges, and applications for deep visual odometry: Toward complicated and emerging areas}.
\newblock \bibinfo{journal}{IEEE Transactions on Cognitive and Developmental Systems} \bibinfo{volume}{14}, \bibinfo{pages}{35--49}.
\bibitem[{Wang et~al.(2017)Wang, Clark, Wen and Trigoni}]{Wang_VO_2017}
\bibinfo{author}{Wang, S.}, \bibinfo{author}{Clark, R.}, \bibinfo{author}{Wen, H.}, \bibinfo{author}{Trigoni, N.}, \bibinfo{year}{2017}.
\newblock \bibinfo{title}{Deepvo: Towards end-to-end visual odometry with deep recurrent convolutional neural networks}, in: \bibinfo{booktitle}{2017 IEEE International Conference on Robotics and Automation (ICRA)}, pp. \bibinfo{pages}{2043--2050}.
\bibitem[{Yang et~al.(2020)Yang, Stumberg, Wang and Cremers}]{Yang_2020_CVPR}
\bibinfo{author}{Yang, N.}, \bibinfo{author}{Stumberg, L.v.}, \bibinfo{author}{Wang, R.}, \bibinfo{author}{Cremers, D.}, \bibinfo{year}{2020}.
\newblock \bibinfo{title}{D3vo: Deep depth, deep pose and deep uncertainty for monocular visual odometry}, in: \bibinfo{booktitle}{Proceedings of the IEEE/CVF Conference on Computer Vision and Pattern Recognition (CVPR)}.
\bibitem[{Zhang et~al.(2025)Zhang, Jiang and Qiu}]{Zhang_Filter_2025}
\bibinfo{author}{Zhang, P.}, \bibinfo{author}{Jiang, C.}, \bibinfo{author}{Qiu, J.}, \bibinfo{year}{2025}.
\newblock \bibinfo{title}{A tightly coupled and invariant filter for visual-inertial-gnss-barometer odometry}.
\newblock \bibinfo{journal}{IEEE Robotics and Automation Letters} \bibinfo{volume}{10}, \bibinfo{pages}{3964--3971}.

\end{thebibliography}

\end{document}